\documentclass[11pt,a4paper]{article}
\usepackage[hyperref]{acl2021}
\usepackage{times}
\usepackage{latexsym}

\usepackage{microtype}
\usepackage{graphicx}
\usepackage{amsmath}
\usepackage{amssymb}
\usepackage{amsfonts}
\usepackage{mathrsfs}
\usepackage{enumitem}
\usepackage{nccmath}
\usepackage{color, colortbl,soul}
\definecolor{LGray}{gray}{0.9}
\definecolor{Gray}{gray}{0.8}
\definecolor{DGray}{gray}{0.7}
\usepackage{hyperref}
\usepackage[hyphenbreaks]{breakurl}
\aclfinalcopy 

\usepackage{adjustbox}
\usepackage{array}

\newcolumntype{R}[2]{%
    >{\adjustbox{angle=#1,lap=\width-(#2)}\bgroup}%
    l%
    <{\egroup}%
}
\newcommand*\rot{\multicolumn{1}{R{55}{.5cm}}}

\title{Evaluation of Thematic Coherence in Microblogs}

\author{Iman Munire Bilal$^1$ \quad Bo Wang$^2$$^,$$^4$ \quad Maria Liakata$^1$$^,$$^3$$^,$$^4$ \\ \textbf{Rob Procter}$^1$$^,$$^4$ \quad \textbf{Adam Tsakalidis}$^3$$^,$$^4$ \\
  $^1$Department of Computer Science, University of Warwick\\
  $^2$Department of Psychiatry, University of Oxford\\
  $^3$School of Electronic Engineering and Computer Science, Queen Mary University of London\\
  $^4$The Alan Turing Institute, London, UK \\
\texttt{\{\tt iman.bilal|rob.procter\}@warwick.ac.uk}\\
\texttt{\{\tt mliakata|bwang|atsakalidis\}@turing.ac.uk}\\
}
\date{}

\begin{document}
\maketitle
\begin{abstract}
Collecting together microblogs representing opinions about the same topics within the same timeframe is useful to a number of different tasks and practitioners. A major question is how to evaluate the quality of such thematic clusters. Here we create a corpus of microblog clusters from three different domains and time windows and define the task of evaluating thematic coherence. We provide annotation guidelines and human annotations of thematic coherence by journalist experts. We subsequently investigate the efficacy of different automated evaluation metrics for the task. We consider a range of metrics including surface level metrics, ones for topic model coherence and text generation metrics (TGMs). While surface level metrics perform well, outperforming topic coherence metrics, they are not as consistent as TGMs. 
TGMs are more reliable than all other metrics considered for capturing thematic coherence in microblog clusters due to being less sensitive to the effect of time windows.
\end{abstract}

\section{Introduction}\label{sec:1}
As social media gains popularity for news tracking, unfolding stories are accompanied by a vast spectrum of reactions from users of social media platforms. Topic modelling and clustering methods have emerged as potential solutions to challenges of filtering and making sense of large volumes of microblog posts  \citep{rosa2011topical,aiello2013sensing,resnik2015beyond,surian2016characterizing}.
Providing a way to access easily a wide range of reactions 
around a topic or event 
has the potential to help those, such as journalists ~\citep{tolmie2017}, police~\citep{procter2013}, health ~\citep{furini2018public} and public safety professionals ~\citep{procter2020}, who increasingly rely on social media to detect and monitor progress of events, public opinion and spread of misinformation.

Recent work on grouping together views about tweets expressing opinions about the same entities has obtained clusters of tweets by leveraging two topic models in a hierarchical approach~\citep{gsdmmlda:1}. 
The theme of such clusters can either be represented by their top-$N$ highest-probability words or measured by the semantic similarity among the tweets. 
One of the questions regarding thematic clusters is how well the posts grouped together relate to each other (\textit{thematic coherence}) and how useful such clusters can be. For example, the clusters can be used to discover topics that have low coverage in traditional news media \citep{zhao2011comparing}. \citet{totemss} employ the centroids of Twitter clusters as the basis for topic specific temporal summaries. 

The aim of our work is to \emph{identify reliable metrics for measuring thematic coherence in clusters of microblog posts}. We define thematic coherence in microblogs as follows: 
Given clusters of posts that represent a subject or event within a broad topic, with enough diversity in the posts to showcase different stances and user opinions related to the subject matter, thematic coherence is the extent to which posts belong together, allowing domain experts to easily extract and summarise stories underpinning the posts. 

To measure thematic coherence of clusters we require robust domain-independent evaluation metrics that correlate highly with human judgement for coherence. A similar requirement is posed by the need to evaluate coherence in topic models. \citet{roder2015exploring} provide a framework for an extensive set of coherence measures all restricted to word-level analysis. \citet{bianchi2020pre} show that adding contextual information to neural topic models improves topic coherence. However, the most commonly used word-level evaluation of topic coherence still ignores the local context of each word. 
Ultimately, the metrics need to achieve an optimal balance between coherence and diversity, such that resulting topics describe a logical exposition of views and beliefs with a low level of duplication. Here we evaluate thematic coherence in microblogs on the basis of topic coherence metrics, while also using research in text generation evaluation to assess semantic similarity and thematic relatedness. 
We consider a range of state-of-the-art text generation metrics (TGMs), such as \emph{BERTScore} ~\citep{bert-score:1}, \emph{MoverScore}~\citep{moverscore} and \emph{BLEURT}~\citep{bleurt}, which we re-purpose for evaluating thematic coherence in microblogs and correlate them with assessments of coherence by journalist experts. The main contributions of this paper are: 
\begin{itemize}
\item We define the task of assessing thematic coherence in microblogs and use it as the basis for creating microblog clusters (Sec.~\ref{sec:3}).\vspace{-.25cm}
\item We provide guidelines for the annotation of thematic coherence in microblog clusters and construct a dataset of clusters annotated for thematic coherence spanning two different domains (political tweets and COVID-19 related tweets). The dataset is annotated by journalist experts and is available \footnote{\url{https://doi.org/10.6084/m9.figshare.14703471}} to the research community (Sec.~\ref{sec:3.5}).\vspace{-.25cm}
\item We compare and contrast state-of-the-art TGMs against standard topic coherence evaluation metrics for thematic coherence evaluation and show that the former are more reliable in distinguishing between thematically coherent and incoherent clusters (Secs~\ref{sec:4},~\ref{sec:5}).
\end{itemize}


\section{Related Work}\label{sec:2}
\textbf{Measures of topic model coherence}: 
The most common approach to evaluating topic model coherence is to identify the latent connection between topic words representing the topic.
Once a function between two words is established, \textbf{topic coherence} can be defined as the (average) sum of the function values over all word pairs in the set of most probable words.~\citet{pmi} use Pointwise Mutual Information (PMI) as the function of choice, employing co-occurrence statistics derived from external corpora.~\citet{mimno} subsequently showed that a modified version of PMI correlates better with expert annotators.~\citet{alsumait2009topic} identified junk topics by measuring the distance between topic distribution and corpus-wide distribution of words.~\citet{coherence_emb} model topic coherence by setting the distance between two topic words to be the cosine similarity of their respective embedded vectors. Due to its generalisability potential we follow this latter approach to topic coherence to measure thematic coherence in tweet clusters. We consider \textit{GloVe}~\cite{glove} and \textit{BERTweet}~\citep{BERTweet} embeddings, derived from language models pre-trained on large external \textit{Twitter} corpora. 
To improve performance and reduce sensitivity to noise, we followed the work of ~\citet{coh_cardinality}, who consider the mean topic coherence over several topic cardinalities $|W| \in \{5, 10, 15, 20 \}$. 

Another approach to topic coherence involves detecting intruder words given a set of topic words, an intruder and a document. If the intruder is identified correctly then the topic is considered coherent. 
Researchers have explored varying the number of `intruders'~\citep{word_intrusion:3} and automating the task of intruder detection~\citep{word_intrusion:2}. 
There is also work on topic diversity \citep{nan-etal-2019-topic}. However, there is a tradeoff between diversity and coherence \citep{wu-etal-2020-short}, meaning  high diversity for topic modelling is likely to be in conflict with thematic coherence, the main focus of the paper. Moreover, we are ensuring semantic diversity of microblog clusters through our sampling strategy (See Sec. \ref{sec:3.4}). 

\vspace{.3cm}
\noindent\textbf{Text Generation Metrics}: \textbf{TGMs} have been of great use in applications such as machine translation ~\citep{moverscore, bert-score:1, meteor2, bleurt}, text summarisation~\citep{moverscore} and image captioning~\citep{cider,bert-score:1,moverscore}, where a machine generated response is evaluated against ground truth data constructed by human experts. Recent advances in contextual language modeling outperform traditionally used \emph{BLEU}~\citep{bleu} and \emph{ROUGE}~\citep{rouge} scores, which rely on surface-level n-gram overlap between the candidate and the reference.

In our work, we hypothesise that metrics based on contextual embeddings can be used as a proxy for microblog cluster thematic coherence. Specifically, we consider the following TGMs:

\noindent\textit{(a) BERTScore} is an automatic evaluation metric based on BERT embeddings~\citep{bert-score:1}. 
The metric is tested for robustness on adversarial paraphrase classification. However, it is based on a greedy approach, where every reference token is linked to the most similar candidate token, leading to a time-performance trade-off. The harmonic mean $F_{BERT}$ is chosen for our task due to its most consistent performance~\citep{bert-score:1}. 

\noindent\textit{(b) MoverScore} ~\citep{moverscore} expands from the \emph{BERTScore} and generalises \emph{Word Mover Distance} ~\citep{word_mover_dist} by allowing soft (many-to-one) alignments. The task of measuring semantic similarity is tackled as an optimisation problem with the constraints given by n-gram weights computed in the corpus. In this paper, we adopt this metric for unigrams and bigrams as the preferred embedding granularity.

\noindent\textit{(c) BLEURT} ~\citep{bleurt} is a state-of-the-art evaluation metric also stemming from the success of BERT embeddings, carefully curated to compensate for problematic training data. Its authors devised a novel pre-training scheme leveraging vast amounts of synthetic data generated through BERT mask-filling, back-translation and word dropping. This allows \emph{BLEURT} to perform robustly in cases of scarce and imbalanced data. 

\section{Methodology}\label{sec:3}

\textbf{Notation} \hspace{.15cm} We use $\mathscr{C}=\{C_1,...,C_n\}$ to denote a set of clusters $C_i$. Each cluster $C_i$ is represented by the pair $C_i=(T_i,W_i)$, where $T_i$ and $W_i$ represent the set of tweets and top-20 topic words of the dominant latent topic in $C_i$, respectively.
\vspace{.1cm}

\noindent\textbf{The task} of identifying thematic coherence in microblog clusters is formalised as follows: Given a set of clusters $\mathscr{C}$, we seek to identify a metric function $f:\mathscr{C}\rightarrow \mathbb{R}$ s.t. high values of $f(C_i)$ correlate with human judgements for thematic coherence. Here we present \textit{(a)} the creation of a corpus of topic clusters of tweets $\mathscr{C}$ and \textit{(b)} the annotation process for thematic coherence. \textit{(a)} involves a clustering (Sec.~\ref{sec:3.2}), a filtering (Sec.~\ref{sec:3.3}) and a sampling step ~(Sec.~\ref{sec:3.4}); \textit{(b)} is described in ~(Sec.~\ref{sec:3.5}). Experiments to identify a suitable function $f$ are in Sec.~\ref{sec:4}.

\subsection{Data Sources}\label{sec:3.1}
We used three datasets pertaining to distinct domains and collected over different time periods as the source of our tweet clusters. 
\vspace{.1cm}

\noindent The \textbf{COVID-19} dataset~\cite{coviddata:1} 
was collected  by tracking COVID-19 related keywords (e.g., \textit{coronavirus}, \textit{pandemic}, \textit{stayathome}) and accounts (e.g., \textit{@CDCemergency}, \textit{@HHSGov}, \textit{@DrTedros}) through the Twitter API from January to May 2020. This dataset covers specific recent events that have generated significant interest and its entries reflect on-going issues and strong public sentiment regarding the current pandemic.

\vspace{.1cm}
\noindent The \textbf{Election} dataset was collected via the Twitter Firehose and originally consisted of all geo-located UK tweets posted between May 2014 and May 2016\footnote{Unlike the Twitter API, the firehose provides 100\% of the tweets that match user defined criteria, which in our case is a set of geo-location and time zone Twitter PowerTrack operators.}. It was then filtered using a list of 438 election-related keywords relevant to 9 popular election issues\footnote{EU and immigration, economy, NHS, education, crime, housing, defense, public spending, environment and energy.} and a list of 71 political party aliases curated by a team of journalists \cite{tdparse}. 

\vspace{.1cm}
\noindent The \textbf{PHEME} dataset~\citep{zubiaga2016} of rumours and non-rumours contains tweet conversation threads consisting of a source tweet and associated replies, covering breaking news pertaining to 9 events (i.e., Charlie Hebdo shooting, German-wings airplane crash, Ferguson unrest, etc.).

These datasets were selected because they cover a wide range of topics garnering diverse sentiments and opinions in the Twitter sphere, capturing newsworthy stories and emerging phenomena of interest to journalists and social scientists. Of particular interest was the availability of stories, comprising groups of tweets, in the PHEME dataset, which is why we consider PHEME tweet clusters separately.

\subsection{Tweet Cluster Generation}\label{sec:3.2}
The task of thematic coherence evaluation introduced in this paper is related to topic modelling evaluation, where it is common practice (~\citet{mimno},~\citet{pmi}) to gauge the coherence level of automatically created groups of topical words. In a similar vein, we evaluate thematic coherence in tweet clusters obtained automatically for the \textbf{Election} and \textbf{COVID-19} datasets. The clusters were created in the following way: Tweets mentioning the same keyword posted within the same time window (3 hours for \textit{Election}, 1 hour for \textit{Covid-19}) were clustered according to the two-stage clustering approach by ~\citet{gsdmmlda:1}, where two topic models ~\citep{gsdmm,lflda} with a tweet pooling step are used. We chose this as it has shown competitive performance over several tweet clustering tasks, without requiring a pre-defined number of clusters. 

The \textbf{PHEME} dataset is structured into conversation threads, where each source tweet is assigned a story label. 
We assume that each story and the corresponding source tweets form a coherent thematic cluster since they have been manually annotated by journalists. Thus the PHEME stories can be used as a gold standard for thematically coherent clusters.
We also created artificial thematically incoherent clusters from PHEME. For this purpose we mixed several stories in different proportions. We designed artificial clusters 
to cover all types of thematic incoherence, namely:~\textit{Random},~\textit{Intruded},~\textit{Chained} (See Sec.~\ref{sec:3.5} for definitions). For \textit{Intruded}, we diluted stories by eliminating a small proportion of their original tweets and introducing a minority of foreign content from other events. For \textit{Chained}, we randomly chose the number of subjects (varying from 2 to 5) to feature in a cluster, chose the number of tweets per subject and then constructed the `chain of subjects' by sampling tweets from a set of randomly chosen stories. Finally, \textit{Random} clusters were generated by sampling tweets from all stories, ensuring no single story represented more than $20\%$ of a cluster.
These 
artificial clusters from PHEME serve as ground-truth data for thematic incoherence. 

\subsection{Cluster Filtering}\label{sec:3.3}
For automatically collected clusters (\textit{COVID-19} and \textit{Election}) we followed a series of filtering steps:  
duplicate tweets, non-English\footnote{\url{https://pypi.org/project/langdetect/}} tweets and ads were removed and only clusters containing 20-50 tweets were kept. 
As we sought to mine stories and associated user stances, opinionated clusters were prioritised. 
The sentiment analysis tool \emph{VADER}~\citep{vader_sentiment} was leveraged to gauge subjectivity in each cluster: a cluster is considered to be opinionated if the majority of its tweets express strong sentiment polarity.\footnote{The absolute value of VADER compound score is required to be $>0.5$, a much stricter condition than that used originally ~\citep{vader_sentiment}.} \emph{VADER} was chosen for its reliability on social media text and for its capacity to assign granulated sentiment valences; this allowed us to readily label millions of tweets and impose our own restrictions to classify neutral/non-neutral instances by varying the thresholds for the VADER compound score. 



\subsection{Cluster Sampling}\label{sec:3.4}
Work on assessing topic coherence operates on either the entire dataset~\citep{word_emb_coh} or a random sample of it \citep{pmi,mimno}. Fully annotating our entire dataset of thematic clusters would be too time-consuming, as the labelling of each data point involves reading dozens of posts rather than a small set of topical words. On the other hand, purely random sampling from the dataset cannot guarantee cluster diversity in terms of different levels of coherence. 
Thus, we opt for a more complex sampling strategy inspired by stratified sampling \citep{stratified}, allowing more control over how the data is partitioned in terms of keywords and scores.

\noindent After filtering \textit{Election} and \textit{COVID-19} contained 46,715 and 5,310 clusters, respectively. We chose to sample 100 clusters from each dataset s.t. they: 
\begin{itemize}
     \item derive from a semantically diverse set of keywords (required for \textit{Elections} only);\vspace{-.28cm}
    \item represent varying levels of coherence (both);\vspace{-.28cm}
    \item represent a range of time periods (both).
\end{itemize}
We randomly subsampled 10 clusters from each keyword with more than 100 clusters 
and keep all clusters with under-represented keywords (associated with fewer than 100 clusters). This resulted in 2k semantically diverse clusters for \textit{Elections}.

TGM scores were leveraged to allow the selection of clusters with diverse levels of thematic coherence in the pre-annotation dataset. 
Potential score ranges for each coherence type were modelled on the PHEME dataset (See Sec.~\ref{sec:3.2},~\ref{sec:3.5}), which is used as a gold standard for cluster coherence/incoherence. For each metric $\mathcal{M}$ and each coherence type $CT$, we defined the associated interval to be:
\begin{equation*}
    I(\mathcal{M})_{CT} = [\mu-2\sigma,\mu+2\sigma],
\end{equation*}
\noindent where $\mu$, $\sigma$ are the mean and standard deviation for the set of metric scores $\mathcal{M}$ characterising clusters of coherence type $CT$. We thus account for 95\% of the data\footnote{Both the \emph{Shapiro-Wilk} and \emph{Anderson-Darling} statistical tests had showed the PHEME data is normally distributed.}.
We did not consider metrics $\mathcal{M}$ for which the overlap between $I(\mathcal{M})_{\text{Good}}$, $I(\mathcal{M})_{\text{Intruded-Chained}}$\footnote{\emph{Intruded} and \emph{Chained} clusters mostly define the intermediate level of coherence, so their score ranges are similar, hence the two groups are unified.} and $I(\mathcal{M})_{\text{Random}}$ was significant as this implied the metric was unreliable. 

As we did not wish to introduce metric bias when sampling the final dataset, we subsampled clusters across the intersection of all suitable metrics for each coherence type $CT$. In essence, our final clusters were sampled from each of the sets  $\mathcal{C}_{CT}= \{C_i | \text{ } \mathcal{M}(C_i)\in I(\mathcal{M})_{CT} \text{ }\forall \text{ metric } \mathcal{M} \}$.
For each of \emph{COVID-19} and \emph{Elections} we sampled 50 clusters $\in \mathcal{C}_\text{Good}$, 25 clusters $\in \mathcal{C}_\text{Intruded-Chained}$ and 25 clusters $\in \mathcal{C}_\text{Random}$.

\subsection{Coherence Annotation Process}\label{sec:3.5}
Coherence annotation was carried out in four stages by three annotators. We chose experienced journalists as they are trained to quickly and reliably identify salient content. 
An initial pilot study including the journalists and the research team was conducted; this involved two rounds of annotation and subsequent discussion to align the team’s understanding of the guidelines (for the guidelines see Appendix B). 

The first stage tackled tweet-level annotation within clusters and drew inspiration from the classic task of word intrusion ~\citep{word_intrusion:1}: annotators were asked to group together tweets discussing a common subject; tweets considered to be `intruders' were assigned to groups of their own. Several such groups can be identified in a cluster depending on the level of coherence. This grouping served as a building block for subsequent stages. This sub-clustering step offers a good trade-off between high annotation costs and manual evaluation since manually creating clusters from thousands of tweets is impractical. We note that agreement between journalists is not evaluated at this first stage as obtaining exact sub-clusters is not our objective. 
However, vast differences in sub-clustering are captured in the next stages in quality judgment and issue identification (See below).
 The second stage concerned cluster quality assessment, which is our primary task. Similar to ~\citet{pmi} for topic words, annotators evaluated tweet cluster  coherence on a 3-point scale (\textit{Good}, \textit{Intermediate}, \textit{Bad}). 
\textit{Good} coherence is assigned to a cluster where the majority of tweets belong to the same theme (sub-cluster), while clusters containing many unrelated themes (sub-clusters) are assigned~\textit{bad} coherence.

The third stage pertains to issue identification of low coherence, similar to ~\citet{mimno}. When 
either \textit{Intermediate} or \textit{Bad} are chosen in stage 2 annotators can select from a list of issues 
to justify their choice: 

\begin{itemize}

\vspace{-0.2cm}\item \textbf{Chained}: several themes are identified in the cluster (with some additional potential random tweets), without clear connection between any two themes.\vspace{-.15cm}
\vspace{-0.2cm}\item \textbf{Intruded}: only one common theme can be identified among some tweets in the cluster and the rest have no clear connection to the theme or to each other.\vspace{-.15cm}
\vspace{-0.2cm}\item \textbf{Random}: no themes can be identified and there is no clear connection among tweets in the cluster.
\end{itemize}
Inter-annotator agreement (IAA) was computed separately for the second and third stages as they serve a different purpose. For the second stage (cluster quality), we obtain average Spearman correlation $r_s=0.73$ which is comparable to previous coherence evaluation scores in topic modelling literature (\citep{pmi} with rs = 0.73 / 0.78 and \citep{aletras} with rs = 0.70 / 0.64 / 0.54) and average Cohen's Kappa $\kappa =0.48$ (moderate IAA). For the third stage (issue identification), we compute average $\kappa = 0.36$ (fair IAA). 

Analysis of pairwise disagreement in stage 2 shows only 2$\%$ is due to division in opinion over Good-Bad clusters. Good-Intermediate and Intermediate-Bad cases account for 37$\%$ and 61$\%$ of disagreements respectively. 
This is encouraging as annotators almost never have polarising views on cluster quality and primarily agree on the coherence of a good cluster, the main goal of this task. For issue identification the majority of disagreements ($\%49$) consists in distinguishing Intermediate-Chained cases. This can be explained by the expected differences in identifying subclusters in the first stage. 
For the adjudication process, we found that a majority always exists and thus the final score was assigned to be the majority label (2/3 annotators).
Table ~\ref{corpus_analysis} presents a summary of the corpus size, coherence quality and issues identified for \textit{COVID-19} and \textit{Election}~(See Appendix C for a discussion).

\begin{table*}[!h]
    \centering
    \resizebox{0.95\textwidth}{!}{
    \begin{tabular}{|l|rrr|rrr|rrr|}\hline
         & \multicolumn{3}{c|}{\textbf{General}} & \multicolumn{3}{c|}{\textbf{Cluster Quality}} & \multicolumn{3}{c|}{\textbf{Cluster Issue}}\\
        \hline
         \textbf{Dataset} & Clusters & Tweets & Tokens & Good & Intermediate & Bad & Intruded & Chained & Random \\
         \textbf{COVID-19}&100&2,955&100K&18&31&51&32&25&25\\
        \textbf{Election}&100&2,650&52K&25&50&25&28&33&14\\
        \hline

    \end{tabular}}
    \caption{Statistics of the annotated clusters where the final label is assigned to be the majority label.}
    \label{corpus_analysis}
\end{table*}

\section{Experiments}\label{sec:4}
Our premise is that a pair of sentences scoring high in terms of TGMs means that the sentences are semantically similar. When this happens across many sentences in a cluster then this denotes good cluster coherence.
~Following \citet{douven2007measuring}, we consider three approaches to implementing and adapting TGMs to the task of measuring thematic coherence. The differences between these methods consist of: (a) the choice of the set of tweet pairs $\mathbb{S}\subset T \times T $ on which we apply the metrics and (b) the score aggregating function $f(C)$ assigning coherence scores to clusters. The TGMs employed in our study are \textit{BERTScore} \citep{bert-score:1}, \textit{MoverScore} ~\citep{moverscore} for both unigrams and bigrams and \textit{BLEURT} \citep{bleurt}. We also employed a surface level metric based on cosine similarity distances between TF-IDF representations\footnote{Tweets are embedded into a vector space of TF-IDF representations within their corresponding cluster.} of tweets to judge the influence of word co-occurrences in coherence analysis.
Each approach has its own advantages and disadvantages, which are outlined below.

\subsection{Exhaustive Approach}\label{sec:4.1}
In this case $\mathbb{S} =$ $T$$\times$$T$, i.e., all possible tweet pairs within the cluster are considered. The cluster is assigned the mean sum over all scores. This approach is not biased towards any tweet pairs, so is able to penalise any tweet that is off-topic. However, it is computationally expensive as it requires $O(|T|^2)$ operations. 
Formally, given a TGM $\mathcal{M}$, we define this approach as:
\begin{equation*}
    f(C)= \dfrac{1}{ \binom{|T|}{2}} \cdot \sum_{\substack{\text{tweet}_i, \text{tweet}_j\in T
\\
i<j}}\mathcal{M}(\text{tweet}_i, \text{tweet}_j).
\end{equation*}

\subsection{Representative Tweet Approach}\label{sec:4.2}
We assume there exists a representative tweet able to summarise the content in the cluster, denoted as the \emph{representative tweet} (i.e. $\text{tweet}_{rep}$). This is formally defined as: 
\begin{equation*}
\mbox{tweet}_{rep}(C)= arg \min_{\substack{\mbox{tweet}_i\in C \\}}D_{KL}(\theta, \mbox{tweet}_{i}),
\end{equation*}
where we compute the \emph{Kullback–Leibler} divergence ($D_{KL}$) between the word distributions of the topic $\theta$ representing the cluster $C$ and each tweet in $C$ \citep{wan2016automatic}; we describe the computation of $D_{KL}$ in Appendix A. 
We also considered other text summarisation methods \citep{basave2014automatic,wan2016automatic} such as \emph{MEAD} ~\citep{centroid} and \emph{Lexrank} ~\citep{lexrank} to extract the best representative tweet, but our initial empirical study indicated $D_{KL}$ consistently finds the most appropriate representative tweet. In this case cluster coherence is defined as below and has linear time complexity $O(|T|)$:
\begin{equation*}
f(C)=\dfrac{1}{|T|} \sum_{\text{tweet}_i\in T}\mathcal{M}(\text{tweet}_i, \text{tweet}_{rep}).
\end{equation*}
As $\mathbb{S} = \{ (\text{tweet},\text{tweet}_{rep})|\text{ tweet}\in T \}\subsetneq T\times T$, the coherence of a cluster is heavily influenced by the correct identification of the representative tweet.

\subsection{Graph Approach}\label{sec:4.3}
Similar to the work of ~\citet{lexrank}, each cluster of tweets $C$ can be viewed as a complete weighted graph with nodes represented by the tweets in the cluster and each edge between tweet$_i$, tweet$_j$ assigned as weight: $w_{i,j}=\mathcal{M}(\text{tweet}_i,\text{tweet}_j)^{-1}.$
In the process of constructing a complete graph, all possible pairs of tweets within the cluster are considered. 
Hence $\mathbb{S} = T \times T$ with time complexity of $O(|T|^2)$ as in Section~\ref{sec:4.1}.
In this case, the coherence of the cluster is computed as the average \textit{closeness centrality} of the associated cluster graph. This is a measure derived from graph theory, indicating how `close' a node is on average to all other nodes; as this definition intuitively corresponds to coherence within graphs, we included it in our study. The closeness centrality for the node representing $\text{tweet}_i$ is given by: 

\begin{equation*} CC(\text{tweet}_i)= \dfrac{|T|-1}{\sum_{\text{tweet}_j\in T} d(\text{tweet}_j,\text{tweet}_i)},\end{equation*}
where $d(\text{tweet}_j,\text{tweet}_i)$ is the shortest distance between nodes tweet$_i$ and tweet$_j$ computed via Dijkstra's algorithm. Note that as Dijkstra's algorithm only allows for non-negative graph weights and \textit{BLEURT}'s values are mostly negative, we did not include this TGM in the graph approach implementation. Here cluster coherence is defined as the average over all closeness centrality scores of the nodes in the graph:\vspace{-.15cm}

\begin{equation*}
    f(C) = \frac{1}{|T|}\sum_{\text{tweet}_\in T}CC(\text{tweet}_i).
\end{equation*}

\section{Results}\label{sec:5}
\begin{table*}[!h]
\centering
\resizebox{.85\textwidth}{!}{%
\begin{tabular}{lccc}
\hline
& \footnotesize{\textbf{Election}} & \footnotesize{\textbf{COVID-19}} & \footnotesize{\textbf{PHEME}}\\
& $r_s$ / $\rho$ / $\tau$ & $r_s$ / $\rho$ / $\tau$ & $r_s$ / $\rho$ / $\tau$ \\
\hline
\footnotesize{\textbf{Exhaustive TF-IDF}} & \footnotesize{\textbf{0.62} / 0.62 / \textbf{0.49}} & \footnotesize{\textbf{0.68} / \textbf{0.72} / \textbf{0.53}} & \footnotesize{0.81 / 0.73 / 0.67 } \\
\footnotesize{\textbf{Graph TF-IDF}} & \footnotesize{\textbf{0.62} / \textbf{0.63} / 0.48} & \footnotesize{0.66 / \textbf{0.72} / 0.52} & \footnotesize{0.74 / 0.71 / 0.60 } \\
\footnotesize{\textbf{Exhaustive BLEURT}} & \footnotesize{0.49 / 0.48 / 0.37} & \footnotesize{0.66 / 0.65 / 0.52} & \footnotesize{\textbf{0.84} / \textbf{0.86} / \textbf{0.69} } \\
\footnotesize{\textbf{Exhaustive BERTScore}} & \footnotesize{0.58 / 0.57 / 0.44} & \footnotesize{0.62 / 0.64 / 0.49} & \footnotesize{0.83 / 0.80 / 0.68} \\
\hline
\footnotesize{\textbf{Topic Coherence Glove}} & \footnotesize{-0.25 / -0.27 / -0.19} & \footnotesize{0.04 / 0.02 / 0.03} & \footnotesize{N/A} \\
\footnotesize{\textbf{Avg Topic Coherence Glove}} & \footnotesize{-0.22 / -0.23 / -0.17} & \footnotesize{-0.03 / -0.03 / -0.02} & \footnotesize{N/A} \\

\footnotesize{\textbf{Topic Coherence BERTweet}} & \footnotesize{-0.23 / -0.22 / -0.18}& \footnotesize{0.10 / 0.11 / 0.08} & \footnotesize{N/A} \\
\footnotesize{\textbf{Avg Topic Coherence BERTweet}} & \footnotesize{-0.17 / -0.16 / -0.14}& \footnotesize{0.04 / 0.04 / 0.03} & \footnotesize{N/A} \\
\hline
\end{tabular}}
\caption{\label{results}
Agreement with annotator ratings across the \textit{Election}, \textit{COVID-19} and \textit{PHEME} datasets. The metrics are Spearman's rank correlation coefficient ($r_s$), Pearson Correlation coefficient ($\rho$) and Kendall Tau ($\tau$).
}
\end{table*}

\normalsize

\subsection{Quantitative Analysis} \label{sec:5.1}
Table~\ref{results} presents the four best and four worst performing metrics (for the full list of metric results refer to Appendix A). \textit{MoverScore} variants are not included in the results discussion as they only achieve average performance. 

\vspace{0.1cm}
\noindent \textbf{Election and COVID-19} \textit{Exhaustive TF-IDF} and \textit{Graph TF-IDF} consistently outperformed TGMs, implying that clusters with a large overlap of words are likely to have received higher coherence scores. While TF-IDF metrics favour surface level co-occurrence and disregard deeper semantic connections, we conclude that, by design all posts in the thematic clusters (posted within a 1h or 3 h window) are likely to use similar vocabulary. 
Nevertheless, TGMs correlate well with human judgement, implying that semantic similarity is a good indicator for thematic coherence: \textit{Exhaustive BERTScore} performs the best of all TGMs in \textit{Election} while \textit{Exhaustive BLEURT} is the strongest competitor to TF-IDF based metrics for \textit{COVID-19}. 

On the low end of the performance scale, we have found topic coherence to be overwhelmingly worse compared to all the TGMs employed in our study. 
\textit{BERTweet} improves over \textit{Glove} embeddings but only slightly as when applied at the word level (for topic coherence) 
it is not able to benefit from the context of individual words. 
We followed ~\citet{coh_cardinality}, and computed average topic coherence across the top $5, 10 , 15, 20$ topical words in order to obtain a more robust performance (see \textit{Avg Topic Coherence Glove}, \textit{Avg Topic Coherence BERTweet}). Results indicate that this smoothing technique correlates better with human judgement for \textit{Election}, but lowers performance further in \textit{COVID-19} clusters.

In terms of the three approaches, we have found that the \textit{Exhaustive} and \textit{Graph} approaches perform similarly to each other and both outperform the \textit{Representative Tweet} approach. Sacrificing time as trade off to quality, the results indicate that metrics considering all possible pairs of tweets account for higher correlation with annotator rankings.

\vspace{0.1cm}
\noindent\textbf{PHEME} The best performance on this dataset is seen with TGM \textit{BLEURT}, followed closely by \textit{BERTScore}. While TF-IDF based metrics are still in the top four, surface level evaluation proves to be less reliable: PHEME stories are no longer constrained by strict time windows\footnote{Stories were generated across several days, rather then hours, by tracking on-going breaking news events on Twitter.}, which allows the tweets within each story to be more lexically diverse, while still maintaining coherence. In such instances, strategies depending exclusively on word frequencies perform inconsistently, which is why metrics employing semantic features (\textit{BLEURT}, \textit{BERTScore}) outperform TF-IDF ones. 
Note that PHEME data lack the topic coherence evaluation, as these clusters were not generated through topic modelling (See \textit{Subsection} \ref{sec:3.2}).

\subsection{Qualitative Analysis}\label{sec:5.2}

\begin{table*}[!h]
\centering
\resizebox{\textwidth}{!}{%
\begin{tabular}{p{\linewidth}}
\rowcolor{Gray} \footnotesize{\textbf{Cluster Annotation: Good}}\hspace{7.8cm}
\footnotesize{\textbf{Common Keyword: `coronavirus'}}\\
\hline

\footnotesize{\textbf{Trump-loving conspiracy nuts tout drinking bleach as a ‘miracle’ cure for coronavirus} - "They may have found a cure for Trump lovers and MAGA but not anything else" $\#$MAGAIDIOTS $\#$TestOnDonJr $\#$OneVoice}\\
\rowcolor{LGray} \footnotesize{Pro-\textbf{Trump conspiracy} theorists \textbf{tout drinking bleach as a 'miracle' cure for coronavirus}}\\
\footnotesize{\textbf{Trump-loving conspiracy nuts tout drinking bleach as a ‘miracle’ cure for coronavirus} – DRINK UP, MAGAts!}\\
\rowcolor{LGray} \footnotesize{Isn't this just a problem solving itself? $\#$Darwinism \textbf{Trump-loving conspiracy nuts tout drinking bleach as a ‘miracle’ cure for coronavirus} }\\
\footnotesize{\textbf{Trump-loving conspiracy nuts tout drinking bleach as a ‘miracle’ cure for coronavirus}...  Is a quart each sufficient? Will go multiple gallons-gratis.}\\
\end{tabular}}
\caption{\label{cluster1}
Cluster fragment from COVID-19 dataset, Exhaustive TF-IDF = 0.109 and Exhaustive BLEURT = -0.808.}
\end{table*}

\begin{table*}[!h]
\centering
\resizebox{\textwidth}{!}{%
\begin{tabular}{p{\linewidth}}
\rowcolor{Gray} \footnotesize{\textbf{Cluster Annotation: Good}} 
\hspace{8cm}\footnotesize{\textbf{Common Keyword: `pandemic'}}\\
\hline
\footnotesize{@CNN @realDonaldTrump administration recently requested $\$$2.5 billion in emergency funds to prepare the U.S. for a possible widespread outbreak of coronavirus. Money isnt necessary if the Trump past 2 years didnt denudate government units that were designed to protect against pandemic}\\
\rowcolor{LGray} \footnotesize{@realDonaldTrump @VP @SecAzar @CDCgov @CDCDirector I bet that was the case of 3 people who had gone no where. You cutting CDC, Scientists $\&$ taking money that was set aside for pandemic viruses that Obama set aside has not helped. You put Pence in charge who did nothing for IN aids epidemic because he said he was a Christian.}\\
\footnotesize{Trump fired almost all the pandemic preparedness team that @BarackObama put in place and his budget promised cutting $\$$ 1.3 billion from @CDC. With 'leadership' like that, what could anyone expect except dire preparation in America?  $\#$ MAGA2020 morons: be careful at his rallies }\\
\rowcolor{LGray} \footnotesize{@USER Democrats DO NOT want mils of Americans to die from coronavirus. They aren't the ones who fired the whole pandemic team Obama put in place. It was Trump. He left us unprepared. All he's interested in is the stock market, wealthy donors $\&$ getting re-elected.}\\
\footnotesize{@USER , Obama set up a pandemic reaction force, placed higher budgets for the CDC AND health and Human Services. Trump on the other hand, have significantly cut the budgets to HHS and the CDC. They disbanded the White House pandemic efforts. With a politician, not a Scientist}\\
\end{tabular}}
\caption{\label{cluster2}
Cluster fragment from COVID-19 dataset, Exhaustive TF-IDF = 0.040 and Exhaustive BLEURT = -0.811.}
\end{table*}

We analysed several thematic clusters to get a better insight into the results. Tables~\ref{cluster1} and \ref{cluster2} show representative fragments from 2 clusters labelled as `good' in the \textit{COVID-19} dataset. The first cluster contains posts discussing the false rumour that bleach is an effective cure to COVID-19, with the majority of users expressing skepticism. As most tweets in this cluster directly quote the rumour and thus share a significant overlap of words, not surprisingly, TF-IDF based scores are high \textit{Exhaustive TF-IDF = 0.109}. In the second cluster, however, users challenge the choices of the American President regarding the government's pandemic reaction: though the general feeling is unanimous in all posts of the second cluster, these tweets employ a more varied vocabulary. Consequently, surface level metrics fail to detect the semantic similarity 
\textit{Exhaustive TF-IDF = 0.040}. When co-occurrence statistics are unreliable, TGMs are more successful for detecting the `common story' diversely expressed in the tweets: in fact, Exhaustive BLEURT assigns similar scores to both clusters (-0.808 for Cluster 1 and -0.811 for Cluster 2) in spite of the vast difference in their content intersection, which shows a more robust evaluation capability.

\begin{table*}[h!]
\centering
\resizebox{\textwidth}{!}{%
\begin{tabular}{p{\linewidth}}
\rowcolor{Gray} \footnotesize{\textbf{Cluster Annotation: Bad Random}}\hspace{8.1cm}
\footnotesize{\textbf{Common Keyword: `oil'}}\\
\hline
\footnotesize{M'gonna have a nap, I feel like I've drank a gallon of like grease or oil or whatever bc I had fish$\&$chips like 20 minutes ago}\\
\rowcolor{LGray} \footnotesize{Check out our beautiful, nostalgic oil canvasses. These stunning images will take you back to a time when life...}\\
\footnotesize{Five years later, bottlenose dolphins are STILL suffering from BP oil disaster in the Gulf. Take action! }\\
\rowcolor{LGray} \footnotesize{Once the gas and oil run out countries like Suadia Arabia and Russia won't be able to get away with half the sh*t they can now}\\
\footnotesize{Ohhh this tea tree oil is burning my face off}\\
\end{tabular}}
\caption{\label{cluster3}
Cluster fragment from Election dataset, TC Glove = 0.330, Exhaustive BERTScore = 0.814 and Exhaustive TF-IDF = 0.024.}
\end{table*}

\begin{table*}[h!]
\centering
\resizebox{\textwidth}{!}{%
\begin{tabular}{p{\linewidth}}
\rowcolor{Gray} \footnotesize{\textbf{Cluster Annotation: Good}}\hspace{8.2cm}
\footnotesize{\textbf{Common Keyword: `migrants'}}\\
\hline
\footnotesize{Up to 300 migrants missing in Mediterranean Sea are feared dead $\#$migrants.}\\
\rowcolor{LGray} \footnotesize{NEWS: More than 300 migrants feared drowned after their overcrowded dinghies sank in the Mediterranean }\\
\footnotesize{Imagine if a ferry sunk with 100s dead - holiday makers, kids etc. Top story everywhere. 300 migrants die at sea and it doesn't lead.}\\
\rowcolor{LGray} \footnotesize{@bbc5live Hi FiveLive: you just reported 300 migrants feared dead. I wondered if you could confirm if the MIGRANTS were also PEOPLE? Cheers.}\\
\footnotesize{If the dinghies were painted pink would there be as much uproar about migrants drowning as the colour of a f**king bus?}\\
\end{tabular}}
\caption{\label{cluster4}
Cluster fragment from Election dataset, TC Glove = 0.307, Exhaustive BERTScore = 0.854 and Exhaustive TF-IDF = 0.100.}
\end{table*}
\vspace{-0.03cm}

We analyse the correlation between topic coherence and annotator judgement in \textit{Tables} \ref{cluster3} and 
\ref{cluster4}. Both are illustrative fragments of clusters extracted from the \textit{Election} dataset. Though all tweets in \textit{Table} \ref{cluster3} share the keyword `oil', they form a bad random cluster type, equivalent to the lowest level of coherence. On the other hand, \textit{Table} \ref{cluster4} clearly presents a good cluster regarding an immigration tragedy at sea. Although this example pair contains clusters on opposite sides of the coherence spectrum, topic coherence metrics fail to distinguish the clear difference in quality between the two. Moreover, \textit{Table} \ref{cluster4} receives lower scores (TC Glove = 0.307) than its incoherent counterpart (TC Glove = 0.330) for Glove Topic Coherence. However, TGM metric BERTScore and surface-level metric TF-IDF correctly evaluate the two clusters by penalising incoherence (Exhaustive BERTScore = 0.814 and Exhaustive TF-IDF = 0.024) and awarding good clusters (Exhaustive BERTScore = 0.854 and Exhaustive TF-IDF = 0.100).
\vspace{-0.1cm}
\section{Conclusions and Future Work}\label{sec:6}
We have defined the task of creating topic-sensitive clusters of microblogs and evaluating their thematic coherence. To this effect we have investigated the efficacy of different metrics both from the topic modelling literature and text generation metrics TGMs.  
We have found that TGMs correlate much better with human judgement of thematic coherence compared to metrics employed in topic model evaluation. TGMs maintain a robust performance across different time windows and are generalisable across several datasets.
In future work we plan to use TGMs in this way to identify thematically coherent clusters on a large scale, to be used in downstream tasks such as multi-document opinion summarisation.

\section*{Acknowledgements}
This work was supported by a UKRI/EPSRC Turing AI Fellowship to Maria Liakata (grant no. EP/V030302/1) and The Alan Turing Institute (grant no. EP/N510129/1).
We would like to thank our 3 annotators for their invaluable expertise in constructing the datasets. We also thank the reviewers for their insightful feedback. Finally, we would like to thank Yanchi Zhang for his help in the redundancy correction step of the pre-processing.

\section*{Ethics}
Ethics approval to collect and to publish extracts from social media datasets was sought and received from Warwick University Humanities \& Social Sciences Research Ethics Committee. During the annotation process, tweet handles, with the except of public figures, organisations and institutions, were anonymised to preserve author privacy rights. In the same manner, when the datasets will be released to the research community, only tweets IDs will be made available along with associated cluster membership and labels.

Compensation rates were agreed with the annotators before the annotation process was launched. Remuneration was fairly paid on an hourly rate at the end of task.

\bibliography{acl2021}
\bibliographystyle{acl_natbib}

\clearpage
\pagenumbering{gobble}
\section*{Appendix A}
\label{sec:appendix}
\setcounter{table}{0}

\subsection*{Representative-Tweet Selection}
\label{appendix:kld}
As described in Section~\ref{sec:4.2}, we select the tweet that has the lowest divergence score to the top topic words of the cluster. Following \citep{wan2016automatic}, we compute the \emph{Kullback–Leibler} divergence ($D_{KL}$) between the word distributions of the topic $\theta$ the cluster $C$ represents and each tweet in $C$ as follows:

\begin{equation*}
\begin{aligned}
&D_{KL}(\theta, \text{\scriptsize{tweet}}_i)\\
&=\sum_{w\in TW\bigcup SW }p_{\theta}(w)*log\frac{p_{\theta}(w)}{tf(w,\text{\scriptsize{tweet}}_i)/len(\text{\scriptsize{tweet}}_i)} 
\end{aligned}
\end{equation*}

where $p_{\theta}(w)$ is the probability of word $w$ in topic $\theta$. $TW$ denotes top 20 words in cluster $C$ according to the probability distribution while $SW$ denotes the set of words in $\text{tweet}_i$ after removing stop words. $tf(w,\text{tweet}_i)$ denotes the frequency of word $w$ in $\text{tweet}_i$, and $len(\text{tweet}_i)$ is the length of $\text{tweet}_i$ after removing stop words. For words that do not appear in $SW$, we set $tf(w,\text{tweet}_i)/len(\text{tweet}_i)$ to 0.00001.

\subsection*{Timings of Evaluation Metrics}
\renewcommand{\thetable}{A\arabic{table}}
\begin{table}[h]
    \centering
    \resizebox{.8\columnwidth}{!}{%
    \begin{tabular}{|c|c|}
         \hline
         \small{\textbf{Metric}} & \small{\textbf{Time (in seconds)}}  \\
         \hline
        \small{Exhaustive BERTScore} & \small{10.84} \\
        \small{Exhaustive BLEURT} & \small{234.10} \\
        \small{Exhaustive MoverScore1} & \small{11.73} \\
        \small{Exhaustive MoverScore2} & \small{11.42} \\
        \small{Exhaustive TF-IDF} & \small{0.05} \\
        \small{Graph TF-IDF} & \small{0.12} \\
        \hline
 
    \end{tabular}}
    \caption{Average timings of metric performances per 1 cluster}
    \label{tab:my_label}
\end{table}

\subsection*{Complete Results}
\label{appendix:metrics}

The complete results of our experiments are in \textit{Table} \ref{full_results}.
The notation is as follows: 
\begin{itemize}
\item \textbf{Exhaustive} indicates that the \textit{Exhaustive Approach} was employed for the metric.
\item \textbf{Linear} indicates that the \textit{Representative Tweet Approach} was employed for the metric.
\item \textbf{Graph} indicates the the \textit{Graph Approach} was employed for the metric.

Shortcuts for the metrics are: \textbf{MoverScore1} = MoverScore applied for unigrams; \textbf{MoverScore2} = MoverScore applied for bigrams

\end{itemize}

\renewcommand{\thetable}{A\arabic{table}}
\begin{table*}[htb]
\centering

\begin{tabular}{lccc}
\hline
& \footnotesize{\textbf{Election}} & \footnotesize{\textbf{COVID-19}} & \footnotesize{\textbf{PHEME}}\\
& $r_s$ / $\rho$ / $\tau$ & $r_s$ / $\rho$ / $\tau$ & $r_s$ / $\rho$ / $\tau$ \\
\hline
\footnotesize{\textbf{Exhaustive TF-IDF}} & \footnotesize{\textbf{0.62} / 0.62 / \textbf{0.49}} & \footnotesize{\textbf{0.68} / \textbf{0.72} / \textbf{0.53}} & \footnotesize{0.81 / 0.73 / 0.67 } \\
\footnotesize{\textbf{Linear TF-IDF}} & \footnotesize{\textbf{0.51} / 0.48 / \textbf{0.39}} & \footnotesize{\textbf{0.36} / \textbf{0.45} / \textbf{0.27}} & \footnotesize{N/A} \\
\footnotesize{\textbf{Graph TF-IDF}} & \footnotesize{\textbf{0.62} / \textbf{0.63} / 0.48} & \footnotesize{0.66 / \textbf{0.72} / 0.52} & \footnotesize{0.74 / 0.71 / 0.60 } \\

\footnotesize{\textbf{Exhaustive BLEURT}} & \footnotesize{0.49 / 0.48 / 0.37} & \footnotesize{0.66 / 0.65 / 0.52} & \footnotesize{\textbf{0.84} / \textbf{0.86} / \textbf{0.69} } \\
\footnotesize{\textbf{Linear BLEURT}} & \footnotesize{0.41 / 0.40 / 0.32} & \footnotesize{0.34 / 0.34 / 0.26} & \footnotesize{N/A} \\

\footnotesize{\textbf{Exhaustive BERTScore}} & \footnotesize{0.58 / 0.57 / 0.44} & \footnotesize{0.62 / 0.64 / 0.49} & \footnotesize{0.83 / 0.80 / 0.68} \\
\footnotesize{\textbf{Linear BERTScore}} & \footnotesize{0.49 / 0.50 / 0.38} & \footnotesize{0.50 / 0.53 / 0.38} & \footnotesize{N/A} \\
\footnotesize{\textbf{Graph BERTScore}} & \footnotesize{0.57 / 0.57 / 0.44} & \footnotesize{0.62 / 0.64 / 0.49} & \footnotesize{0.83 / 0.73 / 0.68}\\

\footnotesize{\textbf{Exhaustive MoverScore1}} & \footnotesize{0.56 / 0.55 / 0.43} & \footnotesize{0.46 / 0.56 / 0.35} & \footnotesize{0.56 / 0.56 / 0.44} \\
\footnotesize{\textbf{Linear MoverScore1}} & \footnotesize{0.54 / 0.52 / 0.41} & \footnotesize{0.36 / 0.39 / 0.28} & \footnotesize{N/A} \\
\footnotesize{\textbf{Graph MoverScore1}} & \footnotesize{0.53 / 0.53 / 0.42} & \footnotesize{0.37 / 0.44 / 0.29} & \footnotesize{0.52 / 0.56 / 0.40}\\

\footnotesize{\textbf{Exhaustive MoverScore2}} & \footnotesize{0.46 / 0.46 / 0.35} & \footnotesize{0.35 / 0.46 / 0.27} & \footnotesize{0.40 / 0.35 / 0.30} \\
\footnotesize{\textbf{Linear MoverScore2}} & \footnotesize{0.47 / 0.46 / 0.35} & \footnotesize{0.26 / 0.31 / 0.20} & \footnotesize{N/A} \\
\footnotesize{\textbf{Graph MoverScore2}} & \footnotesize{0.47 / 0.49 / 0.36} & \footnotesize{0.42 / 0.50 / 0.32} & \footnotesize{0.36 / 0.39 / 0.27}\\

\footnotesize{\textbf{Topic Coherence Glove}} & \footnotesize{-0.25 / -0.27 / -0.19} & \footnotesize{0.04 / 0.02 / 0.03} & \footnotesize{N/A} \\
\footnotesize{\textbf{Avg Topic Coherence Glove}} & \footnotesize{-0.22 / -0.23 / -0.17} & \footnotesize{-0.03 / -0.03 / -0.02} & \footnotesize{N/A} \\

\footnotesize{\textbf{Topic Coherence BERTweet}} & \footnotesize{-0.23 / -0.22 / -0.18}& \footnotesize{0.10 / 0.11 / 0.08} & \footnotesize{N/A} \\
\footnotesize{\textbf{Avg Topic Coherence BERTweet}} & \footnotesize{-0.17 / -0.16 / -0.14}& \footnotesize{0.04 / 0.04 / 0.03} & \footnotesize{N/A} \\

\hline
\end{tabular}
\caption{\label{full_results}
Agreement with human annotators across the \textit{Election}, \textit{COVID-19} and \textit{PHEME} datasets. The metrics are Spearman's rank correlation coefficient ($r_s$), Pearson Correlation coefficient ($\rho$) and Kendall Tau ($\tau$).
}
\end{table*}

\subsection*{PHEME data coherence evaluation}
As original PHEME clusters were manually created by journalists to illustrate specific stories, they are by default coherent. Hence, according to the guidelines, these clusters would be classified as "Good".
For the artificially created clusters, PHEME data is mixed such that different stories are combined in different proportions (See \ref{sec:3.2}). Artificially intruded and chained clusters would be classed as 'Intermediate' as they have been generated on the basis that a clear theme (or themes) can be identified. Finally, an artificially random cluster was created such that there is no theme found in the tweets as they are too diverse; this type of cluster is evaluated as 'Bad'.

\clearpage
\section*{Appendix B: Annotation Guidelines}
\label{appendix:guide}
\vspace{-1cm}
\subsubsection*{Overview}
\vspace{-0.8cm}
You will be shown a succession of clusters of posts from Twitter (tweets), where the posts originate from the same one hour time window. Each cluster has been generated by software that has decided its tweets are variants on the same ‘subject’. You will be asked for your opinion on the quality (‘coherence’) of each cluster as explained below. 
As an indication of coherence quality consider how easy it would be to summarise a cluster.
\vspace{-1.4cm}
\subsubsection*{Steps}
\vspace{-0.7cm}
In the guidelines below, a \textit{subject} is a group of at least three tweets referring to the same topic. 
\textbf{Marking common subjects}: In order to keep track of each subject found in the cluster, label it by entering a number into column \textbf{Subject Label} and then assign the same number for each tweet that you decide is about the same subject. \textbf{Note}, the \textbf{order} of the tweets will automatically change as you enter each number so that those assigned with the same subject number will be listed \textbf{together}.
\vspace{-1.5cm}
\begin{enumerate}
\item 
{\textbf{Reading a Cluster of Tweets}
\vspace{-1.5cm}
\begin{enumerate}
\item Carefully read each tweet in the cluster with a view to uncovering overlapping concepts, events and opinions (if any).
\item Identify the common keyword(s) present in all tweets within the cluster. Note that common keywords across tweets in a cluster are present in all cases by design, so by itself it is not a sufficient criterion to decide on the quality of a cluster.
\item Mark tweets belonging to the same subject as described in the paragraph above.
\end{enumerate}}
\vspace{-2cm}
\item 
{\textbf{Cluster Annotation : What was your opinion about the cluster?}
\vspace{-1.2cm}
\begin{enumerate}
\item Choose ‘\textbf{Good}’ if you can identify one subject within the cluster to which most tweets refer (you can count these based on the numbers you have assigned in the column \textit{Subject Label}). This should be a cluster that you would find it easy to summarise. Proceed to \textbf{Step 4}.
\item Choose ‘\textbf{Intermediate}’ if you are uncertain that the cluster is good, you would find it difficult to summarise its information or you find that there are a small number (e.g., one, two or three) of  unrelated subjects being discussed that are of similar size (\textit{chained}, See issues in Step 3) or one clear subject with a mix of other unrelated tweets (intruded, See issues in Step 3). Additionally, if there is one significantly big subject and one or more other ‘small’ subjects (small ~2,3 tweets), this cluster should be \textit{Intermediate Intruded}. Proceed to \textbf{Step 3}.
\item Choose ‘\textbf{Bad}’ if you are certain that the cluster is not good and the issue of fragmented subjects within the cluster is such that many unrelated subjects are being discussed (heavily \textit{chained}) or there is one subject with a mix of unrelated tweets but the tweets referring to one subject are a minority. Proceed to \textbf{Step 3}. 
\end{enumerate}}

\item
{\textbf{Issue Identification: What was wrong with the cluster?}
\begin{enumerate}
\item Choose ‘\textbf{Chained}’ if several subjects can be identified in the cluster (with some potential random tweets that belong to no subject), but there are no clear connections between any two subjects. This issue can describe both an Intermediate and a Bad cluster.
\item Choose ‘\textbf{Intruded}’ if only one common subject can be identified in some tweets in the cluster and the rest of tweets have no clear connections to the subject or between each other. This issue can describe both an Intermediate and a Bad cluster.
\item Choose ‘\textbf{Random}’ if no subjects can be identified at all as there are no clear connections between the tweets in the cluster. Usually ‘Random’ will be a property of a Bad cluster.
\end{enumerate}}

\item
{\textbf{Cluster Summarisation}
You are asked to provide a brief summary (20-40 words) for each \textbf{Good} cluster you had identified in \textbf{Step 2}. 
}
\end{enumerate}

\clearpage
\section*{Appendix C: Corpus Statistics}
\label{appendix:corpus_stats}

\subsection*{Size}
In terms of size, we observe that the average tweet in \textit{Election} data is significantly shorter (20 tokens) than its correspondent in the \textit{COVID-19} corpus which is 34 tokens long. We observe that the former's collection period finished before Twitter platform doubled its tweet character limit which would be confirmed by the figures in the table. Further work will tackle whether tweet length in a cluster has any impact on the coherence of its message.

\subsection*{Score differences}
We believe differences in the application of the clustering algorithm influenced the score differences between \textit{Election} and \textit{COVID-19} datasets. The clustering algorithm we employed uses a predefined list of keywords that partitions the data into sets of tweets mentioning a common keyword as a first step. The keyword set used for the \textit{Election} dataset contains 438 keywords, while the \textit{COVID-19} dataset contains 80 keywords used for Twitter API tracking~\citep{coviddata:1}. We also note that the different time window span can impact the quality of clusters.

\end{document}